\pdfoutput=1

\documentclass[11pt]{article}

\usepackage[final]{acl}

\usepackage{times}
\usepackage{latexsym}

\usepackage[T1]{fontenc}

\usepackage[utf8]{inputenc}

\usepackage{microtype}

\usepackage{inconsolata}

\usepackage{graphicx}

\newcommand{\cD}{\mathcal{D}}
\newcommand{\cL}{\mathcal{L}}
\newcommand{\cC}{\mathcal{C}}
\newcommand{\cI}{\mathcal{I}}
\newcommand{\cQ}{\mathcal{Q}}
\usepackage{enumitem}
\setlist{leftmargin=2mm}
\usepackage{bbm}
\usepackage{amsmath,amssymb}
\usepackage{multirow}
\usepackage{booktabs}

\title{MoEMoE: Question Guided Dense and Scalable Sparse Mixture-of-Expert for Multi-source Multi-modal Answering}

\author{Vinay Kumar Verma \footnotemark[1] \footnotemark[2] \\
  \small Private Brands - Discovery, Amazon \\
    \texttt{vkvermaa@amazon.com} \And
  Shreyas Sunil Kulkarni \footnotemark[1] \\
  \small International Machine Learning, Amazon \\
  \texttt{kulkshre@amazon.com}
  \AND
  Happy Mittal \footnotemark[2]\\
  \small CMT Systems,  Amazon \\
  \texttt{mithappy@amazon.com} \\
  \And
  Deepak Gupta\\
  \small International Machine Learning,  Amazon \\
  \texttt{dgupt@amazon.com}
}

\begin{document}
\maketitle

\begin{abstract}
Question Answering (QA) and Visual Question Answering (VQA) are well-studied problems in the language and vision domain. One challenging scenario involves multiple sources of information, each of a different modality, where the answer to the question may exist in one or more sources. This scenario contains richer information but is highly complex to handle. In this work, we formulate a novel question-answer generation (QAG) framework in an environment containing multi-source, multimodal information. The answer may belong to any or all sources; therefore, selecting the most prominent answer source or an optimal combination of all sources for a given question is challenging. To address this issue, we propose a question-guided attention mechanism that learns attention across multiple sources and decodes this information for robust and unbiased answer generation. To learn attention within each source, we introduce an explicit alignment between questions and various information sources, which facilitates identifying the most pertinent parts of the source information relative to the question. Scalability in handling diverse questions poses a challenge. We address this by extending our model to a sparse mixture-of-experts (sparse-MoE) framework, enabling it to handle thousands of question types. Experiments on T5 and Flan-T5 using three datasets demonstrate the model's efficacy, supported by ablation studies.

\footnotetext[1]{Equal contribution.}
\footnotetext[2]{This work was done while author was in International Machine Learning team.}
\end{abstract}

\section{Introduction}
\begin{figure}[t]
    \centering
    \includegraphics[scale=0.42]{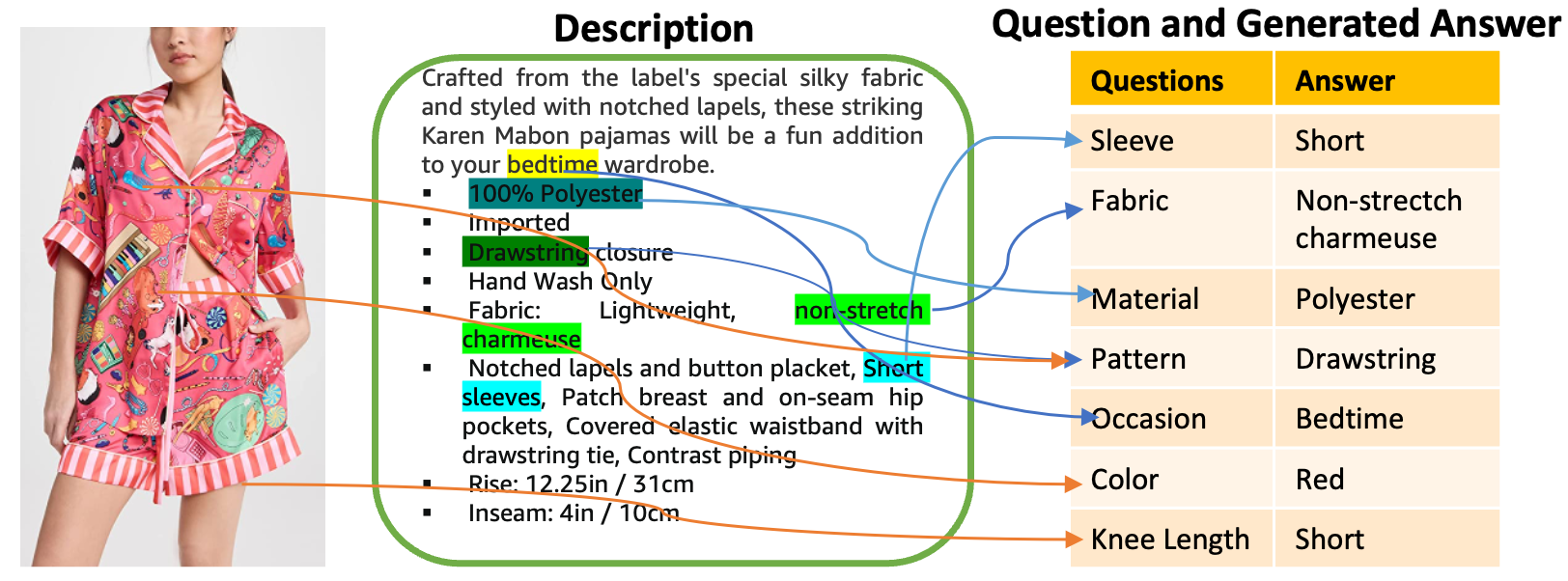}
    \vspace{-7mm}
    \caption{\small Example of the multi-modal and multi-source attribute extraction using the proposed question answering mechanism.} 
    \label{fig:att_example}
    \vspace{-8mm}
\end{figure}

The field of question-answer generation (QAG)~\cite{touvron2023llama,jiang2023mistral} and visual question-answer generation (VQAG)~\cite{li2022blip} holds significant promise with extensive applications across various domains. Recent advancements in large-scale language models~\cite{jiang2023mistral,taori2023alpaca} and vision models~\cite{zhang2023internlm,li2022blip,verma2023skill} have demonstrated notable progress. However, current models are often constrained to QAG tasks that use a single source of information, generating answers solely from either language or visual signals.
In practical applications, handling multiple sources of information is crucial, as answer signals may exist in any or all sources. For example, when reading a paper or article, relying solely on textual content may be insufficient, requiring references to images for a more comprehensive understanding. Similarly, in E-commerce platforms, questions related to attributes such as pattern, fabric, or material can be answered using diverse sources, including images, product descriptions, or external references. Figure~\ref{fig:att_example} illustrates this scenario: questions about attributes like pattern color can be inferred from images, while fabric and material are extracted from text descriptions. Notably, certain attributes, such as pattern, may be present in both sources of information.

Recently, various models~\cite{li2022blip,workshop2022bloom,falcon40b} have emerged for answer generation, leveraging single-source information from either images or text. To handle multi-source information, these models often rely on separate models for different sources, integrating their outputs through post-processing. Novel frameworks, such as PAM~\cite{Lin2021PAMUP} and MXT~\cite{mxt}, have introduced multi-source, multi-modal generative approaches, showing promising results in attribute-related question answering. However, significant challenges remain in developing efficient mechanisms for training and integrating models to handle diverse sources of information.

Existing approaches face several limitations that hinder their effectiveness in answer generation tasks. Firstly, reliance on textual data introduces language bias, potentially leading to skewed attribute generation. Additionally, these models often neglect crucial visual information contained in product images, relying primarily on textual descriptions, which undermines the benefits of multi-source and multi-modal data~\cite{verma2024cod}. Effective answer generation also requires selectively attending to the most relevant source, focusing on key visual or textual information within that source. Furthermore, handling a diverse range of questions with a single model poses scalability challenges, necessitating expert models tailored to specific question types. Unfortunately, models such as MXT~\cite{mxt} and PAM~\cite{Lin2021PAMUP} fail to address these limitations.

\begin{figure*}[ht]
\centering
    \includegraphics[height=5.5cm, width=16cm]{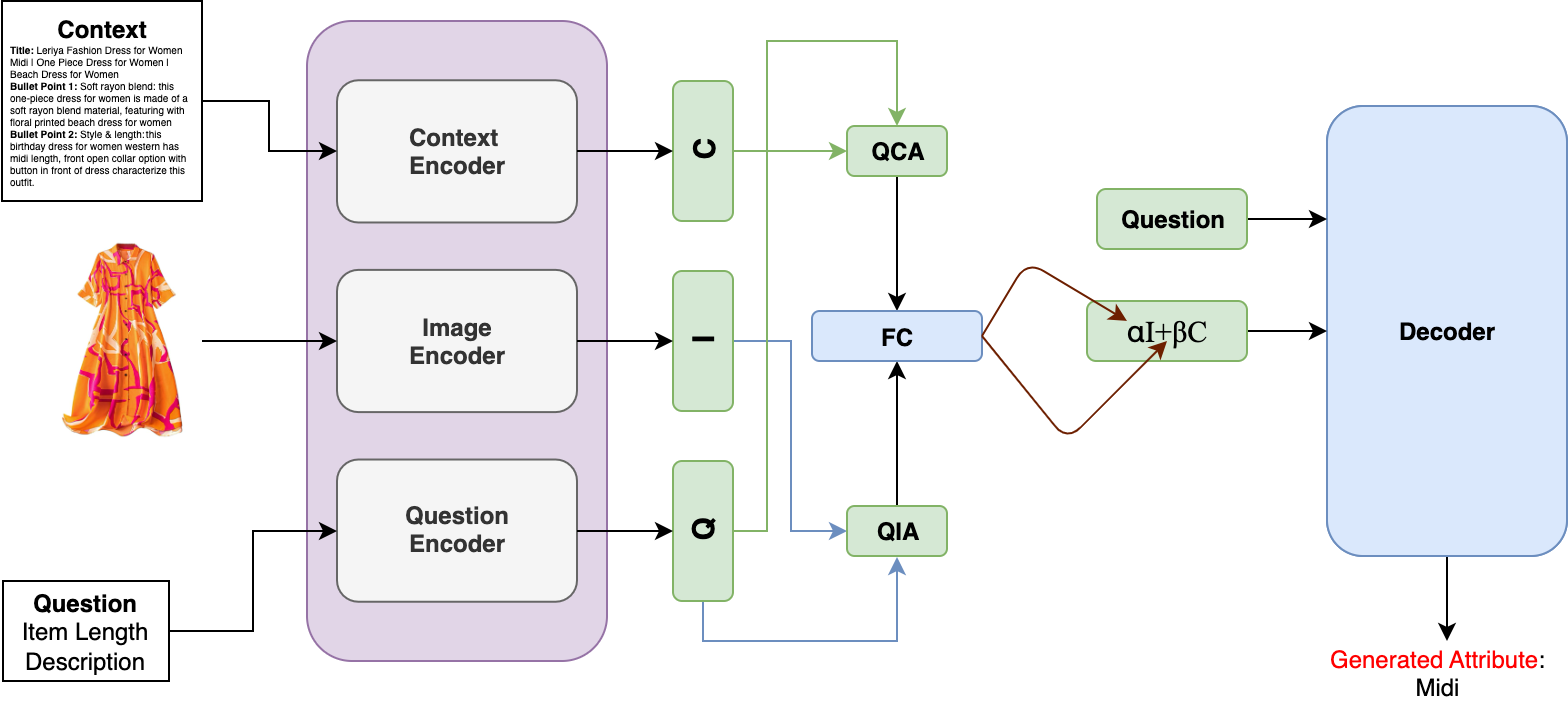}
    \vspace{-9mm}
    \caption{\small The proposed model architecture consists of two T5 encoders for processing the question and context, along with one image encoder. The question and context are aligned using the \emph{Question Context Alignment} Loss while the question and image are aligned through the \emph{Question Image Alignment} Loss}
    \label{fig:proposed_arch}
    \vspace{-6mm}
\end{figure*}

The proposed model addresses the aforementioned limitations by incorporating a question-guided attention (QGA) mechanism and a sparse mixture-of-experts (MoE) model. The QGA mechanism enables the model to autonomously discern attention patterns across multiple sources in scenarios involving diverse information streams. These attention patterns are tailored to the specific posed question. When the answer relies on visual information, the model focuses its attention on visual embeddings. Conversely, when the answer is within the textual context, the model assigns higher weights to textual information. In cases where the answer is derived from all available sources, the model distributes attention appropriately across each source. While cross-modal attention aids in aligning different modalities, it is insufficient for acquiring robust attention patterns within a single source. To address this, we introduce separate embeddings for the question, context, and image, aligning question-image and question-context pairs by maximizing their correlation. This alignment process allows the model to learn precise attention patterns within individual sources based on the given question. Given the diverse nature of the questions, a single model struggles to handle all question types effectively. To address this, We incorporate an MoE strategy into our model, allowing experts to specialize in different question types. Experiments on a large-scale multi-modal dataset show state-of-the-art performance in attribute-based answer generation. Ablation studies analyze the contribution of each model component.
\vspace{-2mm}
\section{Related Work}
\vspace{-2mm}
Extensive work has been conducted on attribute answer extraction, which can be broadly categorized as \emph{extractive}, \emph{predictive}, and \emph{generative}. Extractive models tag each word in a description using Named Entity Recognition (NER) and extract answers based on these tags. Recent works such as OpenTag~\cite{Zheng2018OpenTagOA}, LATEX-numeric~\cite{Mehta2021LATEXNumericLA}, and MQMRC~\cite{Shrimal2022NERMQMRCFN} leverage NER for answer extraction. While effective for certain categories, these models face limitations in predicting novel entities, and defining entity classes remains challenging. Furthermore, NER-based approaches rely solely on unimodal data, ignoring the richer context available in multi-modal sources such as text and images.

Predictive models form another popular category, where answers are predicted from predefined classes using classification models. These approaches accept unimodal or multimodal data with a question and predict attributes from a fixed set. CMA-CLIP~\cite{Liu2021CMACLIPCA} is a recent multimodal approach for attribute prediction. However, these models are limited to predefined attributes and cannot perform zero-shot inference. Given the vast diversity and continuous growth of data, defining a fixed answer set is impractical, and managing large classifier sizes is challenging.

Generative models offer a more flexible solution by generating attributes rather than predicting or extracting them. These models take a question and unimodal or multimodal information as input. AVGPT~\cite{roy2021attribute} generates attribute answers using text data, while PAM~\cite{Lin2021PAMUP} and MXT~\cite{mxt} introduce multimodal generative frameworks. PAM and MXT are closely related to our approach, both employing generative models in multimodal settings. However, MXT uses two image encoders (ResNet152~\cite{He2016DeepRL} and Xception~\cite{chollet2017xception}), making image encoding computationally expensive, and relies on a joint encoder for questions and context. This design prevents direct interaction between the question and the image, limiting the model's ability to focus on relevant image regions. Additionally, MXT uses a cross-modal mechanism that restricts the question's ability to attend to the most relevant source.

Proposed model addresses these limitations by using a single image encoder and learning patch-wise attention, enabling efficient and precise focus on image regions. Furthermore, question-guided attention facilitates attending to the most relevant mode across all sources of information. We also incorporate the MoE~\cite{shazeer2017outrageously}, a recent advancement combining specialized expert models for specific tasks or modalities. MoE has demonstrated significant performance improvements in decoder-only architectures, such as Mixtral~\cite{mixtral} and MoE-LLAVA~\cite{lin2024moellava}, compared to their non-MoE counterparts.
\vspace{-2mm}
\section{Problem Setting}
\vspace{-2mm}
The proposed model solves the QAG task, unlike standard VQA~\cite{mishra2021multi} or QA tasks, our approach incorporates multi-source information, where the answer to a given question may originate from any of the available sources. We define the dataset as $\cD = \{q_i, c_i, i_i\}_{i=1}^N$, comprising $N$ samples, each represented by a triplet of question $q_i$, context $c_i$, and image $i_i$. Here, $q_i$ denotes the question posed for attribute generation, $c_i$ represents the context, including the question, product type (PT), product description, and bullet points, while $i_i$ corresponds to the associated image.
\vspace{-2mm}
\section{Proposed Model}
\vspace{-2mm}
To obtain a robust and highly generalizable model, an approach is needed that can automatically attend to various sources in a multi-modal information scenario. To address this, our approach employs three encoders with unshared parameters for context, image, and question. We have developed a question-guided attention mechanism to automatically learn weights for different data sources. The following section provides a detailed discussion of the proposed model and its components.

\subsection{Source Information Embedding}
Let us consider a context $(c_i)$ and question $(q_i)$, where $c_i, q_i \in \mathbb{R}^{k \times d}$. The context is encoded using the T5 text encoder model with parameters $\theta_c$ and $\theta_q$. Here, $c_i$ includes product descriptions, bullet points, titles, and other relevant information. The T5 architecture is based on the transformer model~\cite{vaswani2017attention} and employs self-attention and multi-head attention (MHA). The encoded embeddings of the context and question are defined as follows: \vspace{-2mm}
\begin{equation} \small \cC = T5_{\theta_c}(c_i), \quad \cQ = T5_{\theta_q}(q_i) : \quad \cQ, \cC \in \mathbb{R}^{k \times d} \label{eq:context} 
\end{equation}
The image is encoded using the SwinV2~\cite{liu2021swin} vision transformer model. Let $i_i$ denote the image, where $i_i \in \mathbb{R}^{3 \times 256 \times 256}$, and let $S$ represent the Swin model with parameter $\theta_s$. The patch embedding from the model is obtained as: \vspace{-4mm}
\begin{align} & i'o = S{\theta_s}(i_i) \quad : \quad q_o \in \mathbb{R}^{k' \times d} \\ & \cI = \text{repeat}(i'_o, \text{int}(k/k')) 
\label{eq:image} 
\end{align} 
Here, we return the patch embedding rather than the final layer logits. The operation $\cI = \text{repeat}(i'_o, \text{int}(k/k'))$ ensures that the image embedding matches the dimensions of the question and context, i.e., dimension $k$.
\vspace{-2mm}
\subsection{Question Guided Attention (QGA)}
The answer to a question may be derived from one or more sources of information. To address this, we developed a QGA mechanism for handling multiple sources. This generic approach applies to any number of sources and can be viewed as a dense MoE, where question-guided attention acts as gating by attending to all sources via a weighted combination. Let $\cQ \in \mathbb{R}^{k \times d}$ represent the question embedding. We transform $q_o$ into a $c$-dimensional embedding using a fully connected (FC) layer, where $c=2$ (representing two sources of information). This operation is given by:\vspace{-4mm}
\begin{equation} 
q_a = FC_{\theta_f}(\cQ), \quad q_a \in \mathbb{R}^{k \times c} 
\label{eq:context} 
\end{equation} 
These $c$-dimensional values for each $k$ are used to assign weights to the various sources. We then learn a joint embedding $e_i$ as follows: 
\begin{equation} 
e_i = \alpha * \cI + \beta * \cC, \quad e_i \in \mathbb{R}^{k \times d} 
\label{eq:image} 
\end{equation} 
Here, $\alpha = q_a[:,0]$ and $\beta = q_a[:,1]$ are $k$-dimensional vectors. Rather than learning a scalar weight for each source, we learn token-specific weights, allowing for finer adjustments compared to a single weight per source.
The joint embedding for the $i^{th}$ sample, $e_i$, is passed to the decoder. Since it is guided by the question, if the answer exists in the context, the model learns a higher weight for the context. If it exists in the image, the image weight is higher. However, for a solution in both sources, the model learns a balanced weight value between sources.

\subsection{Sources and Question Alignment}
In the previous section, the model attends to information from various sources guided by the question but does not learn to focus on relevant information within the source data itself. Here, we align the question with the source data, enabling the model to attend to the most pertinent parts of the source. This alignment is performed for both the image and context relative to the question. The embeddings obtained in Equations \ref{eq:context} and \ref{eq:image} are projected to a single vector of dimension $k$ using a linear transformation and aligned by maximizing cosine similarity. Let $q_p$, $c_p$, and $i_p$ denote the projected embeddings. The alignment losses between the question and sources are defined as:
\begin{align} 
&\cL_{QCA} = \left|1 - (q_p \cdot c_p((|q_p|_2 |c_p|2)\right|, \\
&\cL{QIA} = \left|1 - (q_p \cdot i_p)(|q_p|2 |i_p|2)\right|,
\label{eq:alignment_loss} 
\end{align}
where $\cL{QCA}$ and $\cL{QIA}$ represent the alignment losses for context and image, respectively. This alignment mechanism improves model performance by focusing on the most relevant parts of the source information.
\begin{figure}[ht]
  \centering
    \includegraphics[height=4cm,width=7cm]{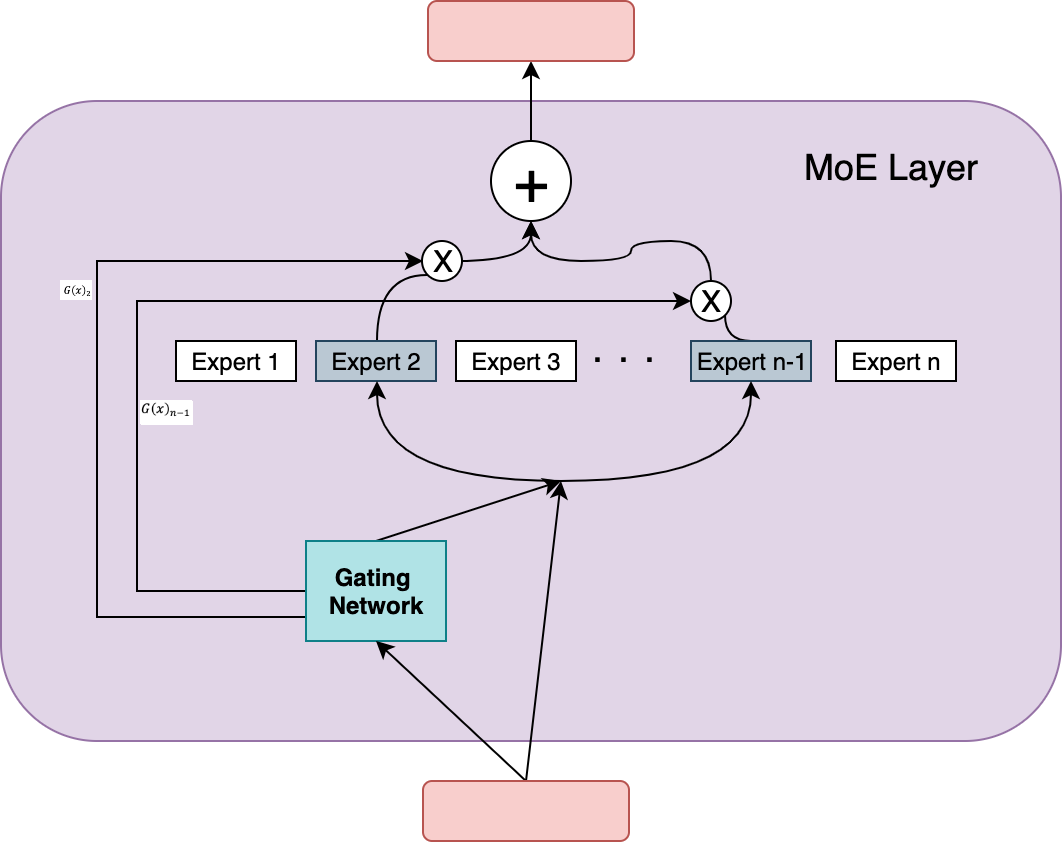}
    \caption{\small Illustration of working of the Mixture of Experts (MoE) Layer \cite{shazeer2017outrageously}}
  \label{fig:moe}
  \vspace{-6mm}
\end{figure}
\subsection{Sparse MoE}
To handle the diverse set of question and obtained a highly scalable model we leverages sparse MoE model (Figure-\ref{fig:moe}), where different expert can handle the various type of questions using a single model. In an MoE framework, we have a set of experts $\{f_1, \ldots, f_n\}$, each taking the same input $x$ and producing outputs $f_1(x), \ldots, f_n(x)$, respectively. Additionally, there is a gating function $w$ that takes $x$ as input and produces a vector of weights $(w(x)_1, \ldots, w(x)_n)$. The gating network is defined by  $w(x) = softmax(top_k(Wx + noise))$. Given an input $x$, the MoE produces a single combined output by aggregating the outputs of the experts $f_1(x), \ldots, f_n(x)$ according to the gating weights $w(x)_1, \ldots, w(x)_n$. At the each layer we choose only $top-k$ expert which produce the sparsity to the model and saves the significant computations. Load balancing is a key issue in the MoE model, to overcome the same we use the load balancing loss. Let $n$ denote the number of experts, and for a given batch of queries $\{x_1, x_2, \dots, x_T\}$, the\textbf{ auxiliary loss} for the batch is defined as: $\cL_{aux}=n\sum_{i=1}^{n} f_i P_i$
Here, $f_i = \frac{1}{T} \#(\text{queries sent to expert }i)$ represents the fraction of times where expert $i$ is ranked highest, and $P_i = \frac{1}{T}\sum_{j=1}^{T}w_i(x_j)$ denotes the fraction of weight assigned to expert $i$, where $w_i(x_j)$ is the weight assigned by the gating mechanism to expert $i$ for query $x_j$.

\subsection{Joint Objective}
Let $g$ is the ground truth token and $\hat{g}$ is the generated token, the decoder loss over the generated token is calculated as follows: $\cL_{\theta_d}(q_i,i_i,c_i)=CrossEntropy(\hat{g}, g)$. The complete objective over the decoder and encoder is given as:
\begin{align}
    \cL_{\theta_q,\theta_i,\theta_c}(q_i,i_i,c_i) & = \cL_{\theta_d}(q_i,i_i,c_i)+\cL_{QCA} \nonumber \\& + \cL_{QIA} +\lambda \cL_{aux}
\end{align}
The model is jointly optimized with respect to parameter $\Theta=[\theta_q,\theta_i,\theta_c,\theta_d, \theta_{f}]$, where $\theta_q$, $\theta_i$, and $\theta_c$ are the encoder parameters, $\theta_f$ is the fully connected layer parameter for question embedding projection, and $\theta_d$ is the decoder parameter.

\begin{table}[t]
\footnotesize  
\begin{center}
    \begin{tabular}{l|c|c|c|c|c}
        \toprule
        \textbf{PT} & \textbf{\#Top} & \textbf{CMA-} & \textbf{NER-} & \textbf{MXT} & \textbf{MoE-}\\
        & \textbf{Attr.} & \textbf{CLIP} & \textbf{MQMRC} & & \textbf{MoE}\\
        \midrule
        \multirow{3}{*}{Kurta} 
        & K=5  & 60.69 & 54.53 & \textbf{76.86} & 76.55\\ 
        & K=10 & 56.67 & 49.97 & 66.86 & \textbf{76.93}\\
        & K=15 & 46.49 & 44.68 & 57.91 & \textbf{60.31}\\
        \midrule
        \multirow{3}{*}{Shirt} 
        & K=5  & 79.60 & 71.26 & 87.89 & \textbf{88.87}\\ 
        & K=10 & 70.47 & 52.01 & 76.99 & \textbf{78.17}\\
        & K=15 & 56.81 & 45.09 & 63.60 & \textbf{69.86}\\
        \bottomrule
    \end{tabular}
\end{center}
\vspace{-4mm}
\caption{\small Results ($Recall@90$) on the 30PT dataset for Kurta and Shirt product types. Results shown for top $K$ attributes ($K=5,10,15$).}
\label{tab:30-pt}
\end{table}

\begin{table}[t!]
\footnotesize
\setlength{\tabcolsep}{4pt}
\begin{center}
\vspace{-2mm}

    \begin{tabular}{l|c|c|c|c|c}
        \toprule
        \textbf{Attribute} & \textbf{CMA-} & \textbf{NER-} & \textbf{KNN} & \textbf{MXT} & \textbf{MoE-}\\
        & \textbf{CLIP} & \textbf{MQMRC} & & & \textbf{MoE}\\
        \midrule
        Color Map & 48.26 & 26.54 & 45.95 & 34.48 & \textbf{49.61}\\ 
        Dress Style & 20.34 & \textbf{20.97} & 13.27 & 23.79 & 20.23\\ 
        Item Length & 66.39 & 47.13 & 63.08 & 65.57 & \textbf{69.92}\\
        Neck & 30.58 & 13.09 & 33.67 & 31.90 & \textbf{34.81}\\
        Pattern & 14.48 & 11.61 & 23.37 & 24.83 & \textbf{41.62}\\
        Season & 67.93 & 16.45 & 65.37 & \textbf{73.10} & 69.43\\
        Sleeve & 61.68 & 35.37 & 44.71 & 54.38 & \textbf{65.77}\\
        \midrule
        \textbf{Average} & 44.23 & 24.45 & 41.34 & 44.01 & \textbf{50.19}\\
        \bottomrule
    \end{tabular}
\end{center}
\vspace{-4mm}
\caption{\small Attribute-wise results ($Recall@90$) on the CMA-CLIP dataset for Dress product type. MoE-MoE outperforms MXT for most attributes, showing significant improvement on average.}
\label{tab:cma-clip}
\vspace{-8mm}
\end{table}
\vspace{-3mm}
\section{Experiment and Results}
\label{sec:experiment}
\vspace{-2mm}
This section, briefly discusses the datasets, baselines and the results obtained using the proposed model.
\vspace{-3mm}

\subsection{Data Description and Base Model}
\vspace{-2mm}
We utilize the \emph{30PT dataset} introduced by MXT~\cite{mxt}, comprising 30 selected product types (PTs) and 38 distinct attributes sourced from an online platform. The CMA-CLIP~\cite{Liu2021CMACLIPCA} paper employed a different dataset with approximately 2.2 million samples for training and 300 samples for validation and testing per attribute. To extend our experiments to a larger scale, we collected a more extensive dataset from the online platform, referred to as OHLSL. This dataset encompasses data for the OHL (other hardlines) and SL (softlines) categories, consisting of 20 million samples for 318 and 145 product types, respectively. The details descriptions about data, baselines and implementations are provided in the supplementary material.
\subsection{Results}
The result over the three standard datasets are discussed below.

\begin{table}[t]
\footnotesize
\centering
\begin{tabular}{l|r|p{0.8cm}|p{0.8cm}|c|p{0.8cm}}
\toprule
\textbf{Attr.} & \textbf{\#Prod} & \textbf{CMA-CLIP} & \textbf{\scriptsize NER-MQMRC} & \textbf{MXT} & \textbf{MoE-MoE} \\
\midrule
age range     & 27.7k & 97.67 & 13.33 & 99.03 & \textbf{99.35} \\
department    & 27.0k & 98.39 & 87.92 & 98.09 & \textbf{98.57} \\
care inst.    & 23.3k & 36.59 & 24.62 & 46.04 & \textbf{48.73} \\
neck    & 22.2k & 52.74 & 48.01 & 68.99 & \textbf{74.47} \\
color   & 21.2k & 84.04 & 74.79 & 86.03 & \textbf{87.65} \\
design  & 19.4k & 24.97 & -- & 32.69 & \textbf{35.41} \\
occas.  & 17.3k & 19.93 & 29.67 & 50.58 & \textbf{52.63} \\
pattern & 13.7k & 25.83 & -- & 31.92 & \textbf{32.61} \\
season  & 12.9k & 5.07 & 0.19 & \textbf{33.20} & 27.15 \\
fit     & 8.0k & 94.66 & 41.59 & 95.61 & \textbf{95.72} \\
closure & 7.7k & 5.05 & -- & 9.33 & \textbf{18.48} \\
collect.& 7.3k & 0.00 & -- & 30.02 & \textbf{46.87} \\
sleeve  & 5.7k & 60.53 & 47.99 & 75.63 & \textbf{80.75} \\
\bottomrule
\end{tabular}
\caption{\small Recall@90P\% on Kurta PT. MoE-MoE shows superior performance on visual attributes (neck, color, design), reducing bias towards textual descriptions.}
\vspace{-4mm}
\label{tab:res30kurta}
\end{table}

\begin{figure}[ht]
	\centering
	\includegraphics[scale=0.3]{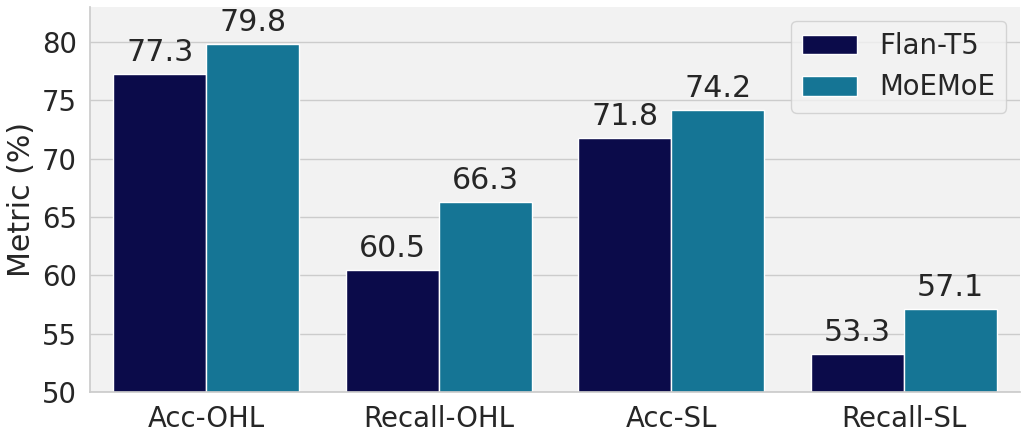}
	\vspace{-5mm}
	\caption{\small Results on the OHLSL dataset over the Flan-T5 architecture. We report the average Accuracy and Recall@90 metric for the OHL and SL category for all the attribute.}
	\label{fig:ohlsl}
	\vspace{-5mm}
\end{figure}
\begin{figure*}[ht]
    \centering
    \includegraphics[scale=0.4]{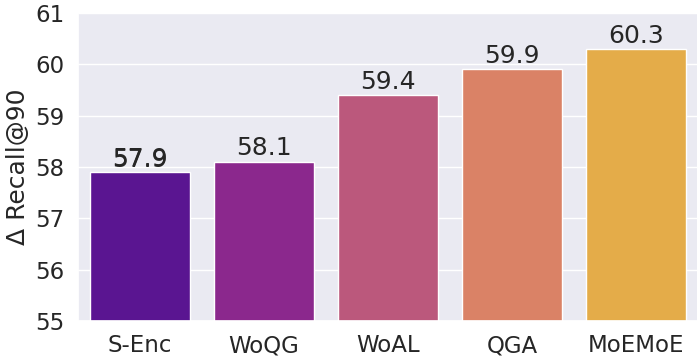}\quad \qquad
    \includegraphics[scale=0.4]{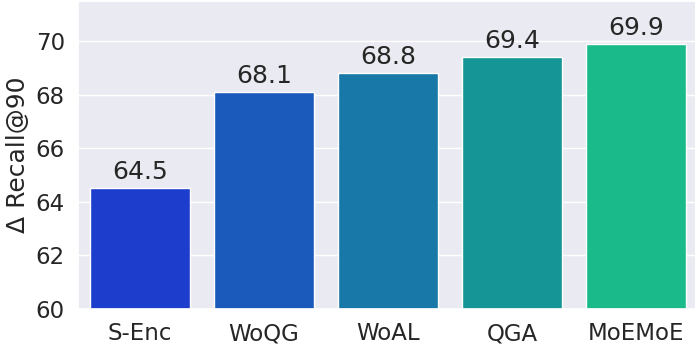}
    \vspace{-3mm}
    \caption{\small The figure shows the ablation over the various component of the proposed model. We can observe that without question guidance (WoQG) or without alignment (WoAL) the model performance significantly drops. Also, single encoder (S-Enc) shows degraded result.}
    \label{fig:ablation}
    \vspace{-6mm}
\end{figure*}
\textbf{30PT dataset}
The 30 PT dataset utilized in our study is a comprehensive dataset containing data from various marketplaces and diverse PTs. Our trained model underwent evaluation on two PTs, namely, \emph{Kurta} and \emph{Shirt}, encompassing $16$ and $19$ attributes, respectively. A detailed breakdown of attribute information is available in the supplementary section. Attributes in our evaluation are associated with either visual information or product descriptions. We employed \emph{Recall@90} (recall with precision $\geq 90$) as our evaluation metric for the top $k$ attributes, where $k=5,10,15$. The results, as presented in Table-\ref{tab:30-pt}, unveil a notable improvement in our proposed approach compared to the recent work MXT. Specifically, our method, MoEMoE, exhibits an absolute improvement of $6.26\%$ and $2.4\%$ over the top 15 attributes for the Shirt and Kurta datasets, respectively. Our analysis indicates that the majority of the improvement over the MXT model stems from attributes related to visual information. In terms of product description-related attributes, both MXT and MoEMoE yield competitive results.

\textbf{CMA-CLIP dataset}
The dataset employed in this study aligns with the one utilized in the CMA-CLIP paper~\cite{Liu2021CMACLIPCA}. Training was conducted using this standardized dataset, and subsequent inference focused on the "dress" category, comprising nine distinct attributes outlined in Table-\ref{tab:cma-clip}. Notably, the proposed model showcases superior performance in six out of the nine attributes when compared to its most competitive counterpart, MXT. Specifically, MoEMoE demonstrates an average absolute improvement of $6.18\%$. In appendix we provide further discussions regarding current model challenges.

\textbf{OHLSL dataset} OHLSL constitutes a large-scale dataset, with each of the OHL and SL categories containing 20 million samples, while the test set encompasses 1 million samples. The results for the OHLSL dataset are evaluated using the Flan-T5 architecture. In Figure-\ref{fig:ohlsl}, the MoEMoE results are compared with those of the Flan-T5 architecture. Notably, we observe an absolute performance improvement of $2.5\%$ and $5.8\%$ for accuracy and recall@90, respectively, over the Flan-T5 architecture in the OHL category. Similarly, for the SL category, we note an absolute improvement of $2.4\%$ and $3.8\%$ over the same architecture.

\vspace{-3mm}
\section{Ablations}
\vspace{-2mm}
We conducted the extensive ablation over the various proposed components and Figure-\ref{fig:ablation} shows the results for the same. Notably, we observed that the presence of question-guided attention and alignment loss had a substantial impact on the model's performance. In the absence of question guidance (WoQG), the performance dropped from $59.9$ to $58.1$ for Kurta PTs and from $69.4$ to $68.1$ for Shirt PTs. When leveraging the joint encoder (S-Enc), where the question and context are merged, the application of question-guided attention was not feasible, leading to a significant performance drop and the lowest results for both Kurta and Shirt product types. The alignment loss emerged as a crucial factor in directing attention within the source information, enabling the model to focus on the most relevant parts of the image or context. Incorporating the alignment loss further enhanced the model's performance, raising it from $59.4$ and $68.8$ to $59.9$ and $69.4$ for the Kurta and Shirt PTs, respectively. The MoE model further helps to improve the model's performance, while maintaining the model's complexity.
\vspace{-2mm}
\subsection{MoE Ablations}
We conducted extensive experiments over various settings discussed in the Section-\ref{sec:experiment}. To the best of our knowledge, there is no existing literature that has conducted experiments for the encoder-decoder architecture. Most recent works on MoE \cite{mixtral} \cite{lin2024moellava} focus on the decoder only architecture. In our experiment we have tried to explore all the experimental scenarios for the encoder-decoder architecture. We measure the results on the Softlines test dataset across 1500 PT-attributes and showcase the results in Table \ref{tab:softlines}. This is a challenging dataset and has a huge, diverse output space across all the product-types.
In our experiments, we investigate the application of the Mixture of Experts (MoE) architecture within the QGA Model over the previously discussed scenarios and our key observations are as follows:

\begin{table}[t]
\vspace{-2mm}
\small
\centering
\caption{\small Results on the Softlines Dataset (1500 PT-attribute test set)}
\vspace{-4mm}
\label{tab:softlines}
\begin{tabular}{p{5cm}|c|c}
\toprule
\textbf{Model} & \textbf{Acc.} & \textbf{R@90} \\ \midrule
MXT  & 63.94 & 53.52 \\
QGA (Question Guided Attention) & 66.04 & 56.69 \\
QGA Enc-Dec MoE Full Training & 62.45 & 54.19 \\
QGA Enc-Dec MoE Odd* & 62.14 & 52.96 \\
QGA Enc-Dec MoE Even* & 62.30 & 53.18 \\
QGA Encoder MoE Full Training & 52.33 & 41.70 \\
QGA Decoder MoE Full Training& 63.81 & 55.08 \\
QGA Decoder Last MoE & 64.29 & 55.70 \\
QGA Decoder Last-2 MoE & 63.67 & 55.07 \\
QGA Decoder Even MoE* & 66.29 & 57.13 \\
QGA Decoder Odd MoE† & 64.79 & 55.50 \\
\textbf{MoEMoE} (\scriptsize QGA Dec. Odd MoE, Expert Training) & \textbf{66.57} & \textbf{57.03} \\ \bottomrule
\multicolumn{3}{@{}l@{}}{\scriptsize *Expert Training Only \quad †MoE Frozen, Backbone Training}
\end{tabular}
\vspace{-6mm}
\end{table}

\textbf{MoE in Decoder}: Applying the MoE architecture exclusively to the decoder layers of the model yields superior performance compared to incorporating it in the encoder layers or across the entire model. This finding suggests that the MoE mechanism is particularly effective in leveraging specialized experts during the output generation phase. Table-\ref{tab:softlines} shows detailed results over the encoder-decoder architecture. The addition of the MoE layer to the full network degrades the model performance and is unable to outperform the base architecture. However, adding the MoE layer to the decoder layer only helps improve the model accuracy and recall@90 by 0.53\% and 0.34\% absolute gain, respectively. However adding the MoE to the encoder-decoder layer degrades the model accuracy and recall@90 by $3.59\%$ and $2.50\%$ respectively in the absolute value. Similarly, adding the MoE to the encoder layer only shows the worst performance and the decrease in the baseline accuracy by $13.71\%$ absolute value. We also observe that training the whole model along with the experts slightly degrades the model performance, however training only the MoE layer helps and outperforms the other baselines. Therefore we can conclude that adding the MoE layer to the decoder only and training the MoE expert only, while freezing the basemodel parameters shows the highest improvement and no other setting works as well. In the future it will be interesting to explore how the internal MoE experts are selected if there are there any intrinsic patterns in the question, context and data that helps to select the MoE expert. In the future we will explore the same.  \\
\textbf{Layer Distribution}: We observe that the choice of applying MoE to even or odd decoder layers does not significantly impact the model's performance, indicating a degree of flexibility in the layer-wise distribution of experts.\\
\textbf{Training Strategy}: The optimal training strategy involves selectively training only the expert modules and the routing network responsible for assigning inputs to experts, while keeping the remaining model parameters frozen. This focused training approach outperforms the conventional end-to-end training of the entire model, including the MoE components. Interestingly, our experiments reveal that fully training the entire model, encompassing the MoE components (experts and routing network) alongside the rest of the model parameters, tends to degrade the overall performance. This observation highlights the potential challenges of jointly optimizing the MoE architecture and the base model in an end-to-end fashion.
\subsection{Auxiliary Loss Ablations}
We conducted the ablation for the MoE loss, the results are shown in the Table \ref{tab:aux-loss}. The ablations are conducted over the best model obtained in the Table-\ref{tab:softlines}. 
\begin{table}[t]
\small
\centering
\caption{\small Impact of Auxiliary Loss Weight on Model Accuracy}
\vspace{-3mm}
\addtolength{\tabcolsep}{5.5mm}
\begin{tabular}{l|c}
\toprule
\textbf{Model Type} & \textbf{Accuracy} \\ \midrule
Enc-Dec (wt=0.01) & 40.34\% \\
Enc-Dec (wt=0.1) & 48.60\% \\
Decoder Only (wt=0.01) & 62.73\% \\
Decoder Only (wt=0.1) & \textbf{66.57\%} \\
Decoder Only (wt=0.5) & 62.95\% \\ \bottomrule
\end{tabular}
\label{tab:aux-loss}
\vspace{-5mm}
\end{table}

We observe that while Enc-Dec MoE with different weights shows degraded results, the decoder-only model demonstrates significant improvement. The MoE loss weight tuning further enhances performance, with $w=0.1$ outperforming other baselines. Too low a weight causes the model to ignore the MoE component, while too high a weight overly prioritizes the MoE loss at the expense of the base model's learning. Thus, the weight must be carefully balanced to enable effective learning of both components.

However, it is important to note that these observations are derived from experiments conducted on a specific task, model architecture, and dataset. The optimal training strategies and deployment of the MoE architecture may vary depending on the problem domain, model characteristics, and data properties.

\section{Conclusions}
\vspace{-2mm}
\label{sec:conclusion}
In this work, we introduce MoEMoE, a robust model designed for question answering from multi-source, multi-modal information. Our approach leverages automatic attention learning across diverse information sources, facilitating the identification of the most reliable source for robust answer generation. The proposed question-guided attention mechanism employs a dense-MoE architecture combined with alignment loss and sparse-MoE training in the intermediate layer, which significantly enhances the model's ability to extract robust features in a scalable manner. The MoEMoE model achieves state-of-the-art results compared to recent baselines. The proposed attention mechanism, operating both between and within multiple sources, is versatile and applicable to various contexts. By incorporating alignment loss between question-context and question-image pairs, the model effectively explores attention within each source, enabling it to focus on the most pertinent parts of the image or context based on the given question. Extensive experiments on a large-scale dataset, coupled with ablation studies, validate the efficacy of our approach.

\newpage
\small
\bibliography{custom.bib}

\newpage
\section{Appendix}
\appendix 
\section{Challenges and Future Work}
\label{sec:limitations}
The proposed approach shows promising results for attribute generation. In spite of its success we face multiple challenges and handling these may further boost the model's performance. These challenges can be summarized as follows:
\paragraph{Pre-trained Tokenizer} The existing model relies on the T5 pre-trained tokenizer, which was trained on a clean and standardized dataset. However, in our specific case, we encounter numerous e-commerce vocabulary terms that are not accounted for in the tokenizer. For instance, the current tokenizer for T5 splits the word "kanchipuram" into tokens such as [$'\_'$, 'kan', 'chi', 'pur', 'am']. Consequently, the tokenized representation fails to capture the intended meaning of the word "kanchipuram." To address this limitation, one potential solution is to train the T5 model using e-commerce data. By incorporating training data specifically tailored to e-commerce, the model can gain a better understanding of the various e-commerce tokens and accurately represent them during tokenization. This approach enables the model to comprehend and handle the unique vocabulary associated with e-commerce domains more effectively.
\paragraph{Mix of English and Non-English} Our current approach utilizes a multi-market dataset; however, it is important to note that the dataset is restricted to the English language. As a result, the model's applicability is limited to English-based marketplaces only. To overcome this constraint and enable deployment in diverse marketplaces, a potential solution lies in incorporating a mixture of both English and Non-English training data. By training the model on a combination of languages, we can expand its language capabilities and ensure its suitability for deployment across various marketplaces.
\paragraph{Hallucination and Out-of-Context Generation}
In future research, there are several intriguing challenges that warrant exploration. One of these challenges pertains to the unrestricted generation of attributes by the model, which occasionally leads to the generation of absurd answers that do not align with the product description or image. This phenomenon, often referred to as "hallucination," involves the model producing answers without factual basis or knowledge. Additionally, there is the issue of questions that cannot be answered accurately based on the given context or image. In such cases, it would be more appropriate for the model to recognize and indicate that the question is out of context, instead of generating a potentially incorrect answer. The challenges of "hallucination" and "Out-of-Context" predictions are noteworthy and present interesting avenues for future exploration. Understanding and addressing these challenges would contribute to the improvement and refinement of attribute generation models.

\section{Dataset Details}
\paragraph{30PT dataset details}
To address the customer problem of accurately identifying attribute values on e-commerce platforms in the IN marketplace, we curated a dataset from a popular e-commerce store. Specifically, we selected 30 product types (PTs) from the apparel category, encompassing a total of 38 unique attributes. Each product in our dataset is accompanied by both text information, including the title, bullet points, and product description, as well as an image. To prepare the data for training, validation, and testing, we applied regex rules to extract and normalize the catalog labels. As an additional step, we performed further normalization by removing infrequently occurring attribute values, accounting for approximately 1\% of the dataset. In terms of the dataset composition, the training data consists of 569k products, while the validation data comprises 84k products, encompassing all 30 PTs. For our test data, we focused on two specific product types: "Shirt" and "Kurta," containing 45k and 28k products, respectively.By leveraging this carefully curated dataset, our research endeavors to improve the overall shopping experience for customers by accurately identifying attribute values on e-commerce platforms in the IN marketplace.

The 30 PTs are: ['apparel belt', 'apparel gloves', 'blazer', 'bra', 'choli', 'coat', 'dress', 'hat', 'kurta', 'necktie', 'nightgown nightshirt', 'overalls', 'pajamas', 'pants', 'saree', 'scarf', 'shirt', 'shorts', 'skirt', 'sleepwear', 'socks', 'suit', 'sweater', 'sweatshirt', 'swimwear', 'tights', 'track suit', 'tunic', 'undergarment slip', 'underpants']. 

The 38 unique attributes are: ['age range description', 'belt style', 'bottom style', 'care instructions', 'closure type', 'collar style', 'collection', 'color', 'cup size', 'department', 'design name', 'fabric type', 'fabric wash', 'finish type', 'fit type', 'front style', 'item length description', 'item styling', 'lining description', 'material', 'neck style', 'occasion type', 'outer material', 'pattern type', 'pocket description', 'rise style', 'seasons', 'sleeve type', 'sport type', 'strap type', 'style', 'subject character', 'theme', 'toe style', 'top style', 'underwire type', 'waist style', 'weave type'].

\section{Data Description and Base Model}
We utilize the \emph{30PT dataset} introduced by MXT~\cite{mxt}, comprising 30 selected product types (PTs) and 38 distinct attributes sourced from an online platform. Each product within this dataset is characterized by its description, bullet points, title, and image. The textual information, encompassing description, bullet points, and title, is amalgamated and treated as context, while the image is considered a separate source of information. Additionally, the marketplace and PT details are concatenated into the context, functioning as a prior for the product type. The model's task is to generate answers for attribute-based questions, treating the attribute name as the question prompt. The dataset includes 5.69 million and 0.84 million products in the training and validation sets, respectively, across 30 PTs. The test data consists of products from two product types, totaling 0.73 million products. The CMA-CLIP~\cite{Liu2021CMACLIPCA} paper employed a different dataset with approximately 2.2 million samples for training and 300 samples for validation and testing per attribute. Although this dataset is simpler compared to the MXT dataset, we incorporate it to assess the robustness of our proposed model across diverse datasets. Both datasets mentioned above have relatively small training and testing sets. To extend our experiments to a larger scale, we collected a more extensive dataset from the online platform, referred to as OHLSL. This dataset encompasses data for the OHL (other hardlines) and SL (softlines) categories, consisting of 20 million samples for 318 and 145 product types, respectively. It includes a total of 197 attribute types for OHL and 108 attribute types for SL.

The proposed model is versatile and applicable to any combination of image encoder and encoder-decoder language model. To assess its generalization ability, we implemented our approach using two widely used language models, T5 and Flan-T5. Our observations indicate a substantial improvement compared to the baseline in both cases. Additionally, we noticed that the Swin~\cite{liu2021swin} image encoder outperforms the ViT model, although the difference is not statistically significant. Consequently, throughout the paper, we consistently employ the Swin image encoder.

\section{Implementation Details}
In our proposed model, we employ the Swin Transformer as the image encoder and T5/Flan-T5 as the language encoder. The question and context undergo separate encoding processes, with no parameter sharing between them. Both the Swin and T5 architectures utilize the base configuration, with 20 million and 250 million parameters, respectively. Training spans 10 epochs for each dataset, utilizing the Adam optimizer with a step learning rate. The initial learning rate is set at $0.001$ and is decayed by a factor of 0.2 at the 6th and 9th epochs. Notably, we assign equal weight of one to all loss functions, eliminating the need for hyperparameter tuning in loss weights. To ensure a fair comparison, we maintain consistency in optimizer choice and training schedule across all models in the baselines. Training employs the distributed data parallel API on an 8xA10 machine with 24GB of memory, utilizing a batch size of four per GPU.  

\section{Baselines and Metrics}
We include recent generative and predictive models as our baselines. For the predictive model, our choice is NER-MQMRC~\cite{Shrimal2022NERMQMRCFN}, an approach based on named entity recognition that demonstrates promising outcomes, particularly in within-class scenarios. As for generative models, we incorporate CMA-CLIP~\cite{Liu2021CMACLIPCA} and MXT~\cite{mxt}. Both of these models are multi-modal approaches and showcase state-of-the-art performance in attribute answer generation through a question-answering approach. CMA-CLIP and MXT follow similar settings, being trained with multi-modal and multi-source information for question-answering. To ensure a fair comparison with these approaches, we refrain from introducing any additional information into our data. 

\section{MoE Detailed}
In an MoE framework, we have a set of experts $\{f_1, \ldots, f_n\}$, each taking the same input $x$ and producing outputs $f_1(x), \ldots, f_n(x)$, respectively. Additionally, there is a gating function $w$ that takes $x$ as input and produces a vector of weights $(w(x)_1, \ldots, w(x)_n)$. The gating network is defined by  $w(x) = softmax(top_k(Wx + noise))$. 


Given an input $x$, the MoE produces a single combined output by aggregating the outputs of the experts $f_1(x), \ldots, f_n(x)$ according to the gating weights $w(x)_1, \ldots, w(x)_n$. The precise aggregation mechanism varies across different MoE implementations. Both the experts and the gating function are trained jointly by minimizing a chosen loss function, typically through gradient-based optimization methods.

Vanilla Mixture of Experts (MoE) models often encounter load balancing issues, where some experts are consulted more frequently than others, leading to an imbalanced utilization of the available experts. To mitigate this issue and encourage the gating mechanism to select each expert with equal frequency within a batch, thereby achieving proper load balancing, we introduce an auxiliary loss function for each MoE layer.

Let $n$ denote the number of experts, and for a given batch of queries $\{x_1, x_2, \dots, x_T\}$, the\textbf{ auxiliary loss} for the batch is defined as:

\begin{equation}
\cL_{aux}=n\sum_{i=1}^{n} f_i P_i
\end{equation}

Here, $f_i = \frac{1}{T} \#(\text{queries sent to expert }i)$ represents the fraction of times where expert $i$ is ranked highest, and $P_i = \frac{1}{T}\sum_{j=1}^{T}w_i(x_j)$ denotes the fraction of weight assigned to expert $i$, where $w_i(x_j)$ is the weight assigned by the gating mechanism to expert $i$ for query $x_j$.

This auxiliary loss is minimized at $1$, precisely when every expert has an equal weight of $\frac{1}{n}$ in all situations, thereby achieving uniform expert utilization within the batch. By incorporating this auxiliary loss into the overall training objective, the gating mechanism is encouraged to distribute the queries evenly across the available experts, mitigating the load balancing issues commonly observed in vanilla MoE models.

Figure 3 in the main paper shows the MoE Layer Diagrammatically
While MoE models have been successfully applied to decoder-only architectures, our study represents the first attempt to integrate MoE into an encoder-decoder framework for multimodal attribute extraction.

By leveraging the Mixture of Experts (MoE) concept, we hypothesize that our proposed model will capture the complex relationships between different modalities more effectively, leading to improved attribute extraction performance. The expert sub-networks can handle multiple attributes efficiently, as similar attributes can activate related experts, allowing different experts to specialize in distinct attribute types. This contrasts with traditional models, where all attributes must be processed by the same layers. Furthermore, MoE's parameter efficiency enables scaling our model to larger capacities without significantly increasing computational cost during inference, ensuring practical applicability.

Apart from changing the general architecture we also have to tweak the loss function. In addition to the traditional logit loss, and the losses from QGA obtained from the question image alignment loss, and question context alignment loss, we augment the overall objective function with the auxiliary loss obtained from the MoE layers. The introduction of the auxiliary loss from the MoE module serves as a regularization mechanism, guiding the model to learn a more diverse and robust set of representations and encouraging the effective distribution of computational load across the experts, thereby enhancing its ability to capture and leverage specialized knowledge for the given task.

\begin{align}
    & \cL_{\theta_q,\theta_i,\theta_c}(q_i,i_i,c_i) \\ &=\cL_{\theta_d} \nonumber(q_i,i_i,c_i)+\cL_{QCA}+\cL_{QIA}+w*\cL_{aux}
\end{align}

The model is jointly optimized with respect to parameter $\Theta=[\theta_q,\theta_i,\theta_c,\theta_d, \theta_{f}]$, where $\theta_q$, $\theta_i$, and $\theta_c$ are the encoder parameters, $\theta_f$ is the fully connected layer parameter for question embedding projection, and $\theta_d$ is the decoder parameter.
Equation $2$ defines the loss function of the MoE-MoE model. $\cL_{\theta_d}(q_i,i_i,c_i)$ stands for the general Cross Entropy loss, $\cL_{QCA}$ stands for the Question-Context Alignment loss, $\cL_{QIA}$, represents Question-Image Alignment loss and finally the $\cL_{aux}$ represents the Auxiliary Loss from the MoE layers multiplied by $w$ which is a weight that we multiply $\cL_{aux}$ by to determine how much importance we give to the Experts and gating network.

\end{document}